\documentclass[runningheads]{llncs}
\usepackage{graphicx}
\usepackage{amsmath,amssymb} 
\usepackage{color}
\usepackage[width=122mm,left=12mm,paperwidth=146mm,height=193mm,top=12mm,paperheight=217mm]{geometry}
\usepackage{times}
\usepackage{epsfig}
\usepackage{graphicx}
\usepackage{amsmath}
\usepackage{amssymb}
\usepackage{amssymb}
\usepackage{multirow}
\usepackage{array}
\usepackage{booktabs}
\usepackage{tabularx}
\usepackage{algorithm}
\usepackage{algorithmic}
\usepackage{stmaryrd}
\usepackage{color}
\usepackage{makecell}
\usepackage{subfigure}

\begin{document}
\pagestyle{headings}
\mainmatter
\def\ECCV18SubNumber{559}  

\title{CIRL: Controllable Imitative Reinforcement Learning for Vision-based Self-driving} 



\authorrunning{X. Liang, T. Wang, L. Yang, E. Xing}

\author{Xiaodan Liang$^{1,2}$, Tairui Wang$^{2}$, Luona Yang$^{1}$, Eric P. Xing$^{2}$}

\institute{Carnegie Mellon University$^{1}$ \quad Petuum Inc.$^{2}$}

\maketitle

\begin{abstract}
Autonomous urban driving navigation with complex multi-agent dynamics is under-explored due to the difficulty of learning an optimal driving policy. The traditional modular pipeline heavily relies on hand-designed rules and the pre-processing perception system while the supervised learning-based models are limited by the accessibility of extensive human experience. We present a general and principled Controllable Imitative Reinforcement Learning (CIRL) approach which successfully makes the driving agent achieve higher success rates based on only vision inputs in a high-fidelity car simulator. To alleviate the low exploration efficiency for large continuous action space that often prohibits the use of classical RL on challenging real tasks, our CIRL explores over a reasonably constrained action space guided by encoded experiences that imitate human demonstrations,  building upon Deep Deterministic Policy Gradient (DDPG). Moreover, we propose to specialize adaptive policies and steering-angle reward designs for different control signals (i.e. follow, straight, turn right, turn left) based on the shared representations to improve the model capability in tackling with diverse cases. Extensive experiments on CARLA driving benchmark demonstrate that CIRL substantially outperforms all previous methods in terms of the percentage of successfully completed episodes on a variety of goal-directed driving tasks. We also show its superior generalization capability in unseen environments. To our knowledge, this is the first successful case of the learned driving policy by reinforcement learning in the high-fidelity simulator, which performs better than supervised imitation learning.



\end{abstract}

\section{Introduction}

Autonomous urban driving is a long-studied and still under-explored task~\cite{pomerleau1989alvinn,silver2010learning} particularly in the crowded urban
environments~\cite{paden2016survey}. A desirable system is required to be capable of solving all visual perception tasks (e.g. object and lane localization, drivable paths) and determining long-term driving strategies, referred as ``driving policy". Although visual perception tasks have been well studied by resorting to supervised learning on large-scale datasets~\cite{zhang2016faster,liang2018dynamic}, simplistic driving policies by manually designed rules in the modular pipeline is far from sufficient for handling diverse real-world cases as discussed in~\cite{shalev2016safe,sallab2017deep}. Learning a optimal driving policy that mimics human drivers is less explored but key to navigate in complex environments that requires understanding of multi-agent dynamics, prescriptive traffic rule, negotiation skills for taking left and right turns, and unstructured roadways. These challenges naturally lead people to machine learning approaches for discovering rich and robust planning strategies automatically.

A line of researches~\cite{bojarski2016end,xu2017end,kim2017interpretable,codevilla2017end,muller2006off,hou2017fast} for learning policies follow the end-to-end imitation learning that directly maps sensor inputs to vehicle control commands via supervised training on large amounts of human driving data. However, these systems cannot be generalized to unseen scenarios and their performances are severely limited by the coverage of human driving data. For example, the model of Bojarski et al.~\cite{bojarski2016end} trained for road following fails for turning right/left. Moreover, it is difficult
to pose autonomous driving with long-term goal-oriented navigation as a supervised learning problem as the autonomous vehicle needs to heavily interact with the environment including other vehicles, pedestrians and roadways.

It is thus desirable to have a richer control policy which considers a large amount of feedbacks from the environment including self-states, collisions and off-road conditions for autonomous driving. Deep reinforcement
Learning (RL) offers, in principle, a reasonable system to learn such policies from exploration~\cite{sutton1998reinforcement}. However, the amount of exploration required for large action space (such as a sequence of continuous steer angles, brakes and speeds) has prohibited its use in real applications, leading to unsatisfactory results by recent efforts on RL-based driving policy learning~\cite{dosovitskiy2017carla,shalev2016safe} in complex real-world tasks.

\begin{figure*}[!tp]
        \begin{center}
            \includegraphics[scale=0.35]{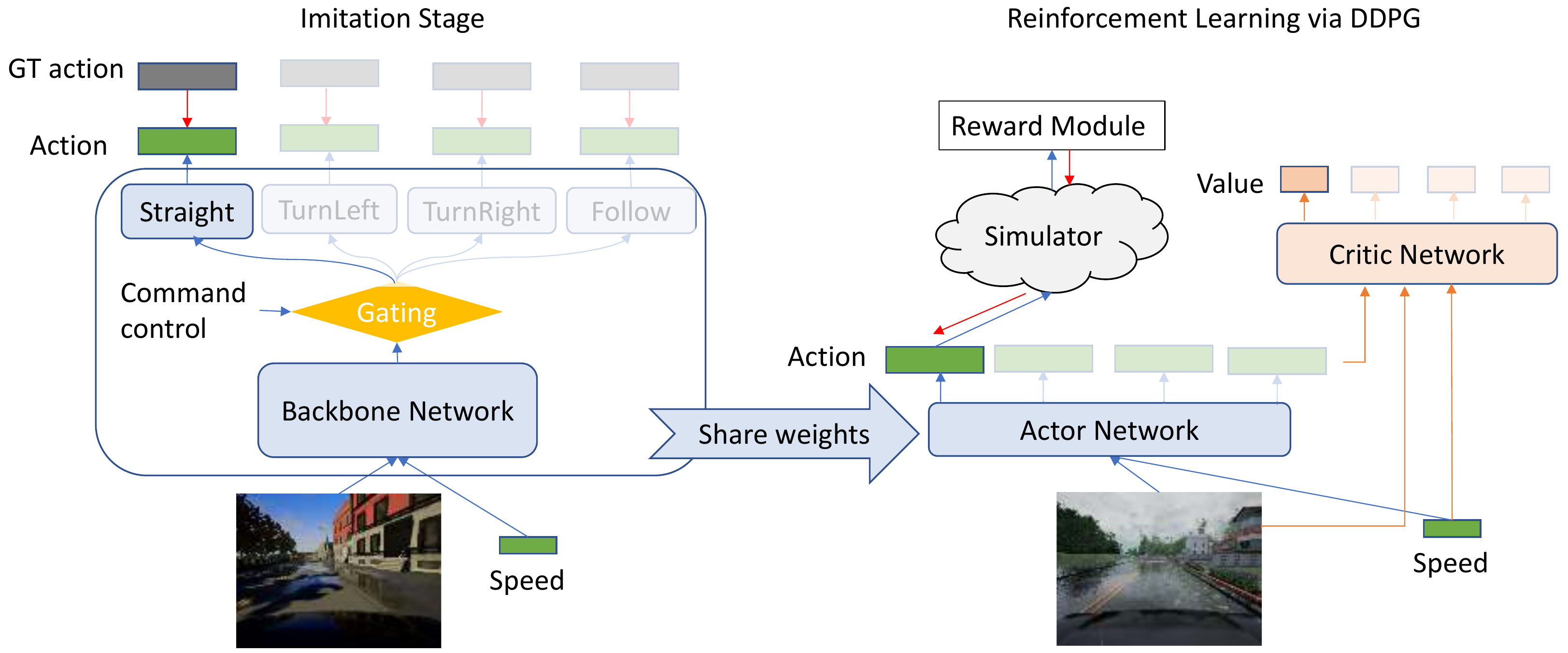}\vspace{-3mm}
            \caption{An overview of our Controllable Imitative Reinforcement Learning (CIRL), including a controllable imitation stage and a reinforcement learning stage optimized via Deep Deterministic Policy Gradient (DDPG). The imitation stage first train the network by supervised learning with groundtruth actions from recorded human driving videos. Then we share the learned weights into the actor network and optimize both actor and critic with feedbacks from reward module by interacting with the simulator.} \vspace{-8mm}
            \label{fig:framework}
        \end{center}
    \end{figure*}
    
In this paper, we resolve this challenging planning task with our novel Controllable Imitative Reinforcement Learning (CIRL) that facilitates the continuous controllable deep-RL by exploiting the knowledge learned from demonstrations of human experts. The whole architecture is illustrated in Fig.~\ref{fig:framework}. Our CIRL is based on the Deep Deterministic Policy Gradient (DDPG)~\cite{lillicrap2015continuous} that is an off-policy replay-memory-based actor-critic algorithm. The conventional DDPG often falls into local optimal due to too much failed explorations for large action space. Our CIRL solves this issue by providing better exploration seeds for the search over the action space of the actor networks. Specifically, the actor networks are first warmed up by learned knowledge via imitation learning using human demonstrations in order to initialize the action exploration in a reasonable space. Then our CIRL incorporates DDPG to gradually boost the generalization capability of the learned driving policy guided by continuous reward signals sent back from the environment. 
Furthermore, to support the goal-oriented navigation, we introduce a controllable gating mechanism to selectively activate different branches for four distinct control signals (i.e. follow, straight, turn right, turn left). Such gating mechanism not only allows the model to be controllable by a central planner or the drivers' intent, but also enhances the model's capability by providing tailored policy functions and reward designs for each command case. In addition, distinct abnormal steer angle rewards are further proposed to better guide policies of each control signal as auxiliary aggregated rewards. 

Our key \textbf{contributions} can be summarized as: 1) we present the first successful deep-RL pipeline for vision-based autonomous driving that outperforms previous modular pipeline and other imitation learning on diverse driving tasks on the high-fidelity CARLA benchmark; 2) we propose a novel controllable imitative reinforcement learning approach that effectively alleviates the inefficient exploration of large-scale continuous action space; 3) a controllable gating mechanism is introduced to allow models be controllable and learn specialized policies for each control signal with the guidance of distinct abnormal steer-angle rewards; 4) comprehensive results on public CARLA benchmark demonstrates our CIRL achieves state-of-the-art performance on a variety of driving scenarios and superior generalization capability by applying the same agent into unseen environments. More successfully driving videos are presented in Supplementary.

\section{Related Work}

Autonomous driving has recently attracted extensive research interests~\cite{paden2016survey}. In general, prior approaches can be categorized into two different pipelines based on the modularity level. The first type is the  highly tuned system that assembles a bunch of visual perception algorithms and then uses model-based planning and control~\cite{book}. Recently, more efforts have been devoted to the second type, that is, end-to-end approaches that learn to map sensory input to
control commands~\cite{xu2017end,bojarski2016end,pomerleau1989alvinn,zhang2017query,codevilla2017end,yang2018unsupervised}. Our method belongs to the second spectrum. 

\textbf{End-to-end Supervised Learning.} The key to autonomous driving is the ability of learn driving policy that automatically outputs control signals for steering wheel, throttle, brake, etc., based on observations. As a straight-forward idea, imitation learning that learns policies via supervised training on human driving data has been applied to a variety of tasks,
including modeling navigational behavior~\cite{ziebart2008navigate}, off-road driving~\cite{muller2006off,silver2010learning}, and road following~\cite{xu2017end,bojarski2016end,pomerleau1989alvinn,zhang2017query,codevilla2017end}. These works differ in several aspects: the input representation (raw sensory input or pre-processed signals), predicting distinct control signals, experimenting on simulated or real data. Among them, ~\cite{pomerleau1989alvinn,muller2006off,codevilla2017end,bojarski2016end} also investigated training networks for directly mapping vision inputs into control signals. The very recent work~\cite{codevilla2017end} relates to our CIRL in incorporating control signals into networks. However, supervised approaches usually require a
large amount of data to train a model that can generalize to different environments. Obtaining massive data for all cities, scenarios and dynamical requires significant human involvement and is impractical since we cannot cover all possible situations that may happen. From the technical aspect, different from these works, our CIRL aims to learn advanced policies by interacting with the simulator guided by the imitation learning towards more and general complex urban driving scenarios. In addition, distinct abnormal steer-angle rewards are defined for each control signal, enabling the
model to learn coherent specialized policies with human commonsense.

\begin{figure*}[!tp]
        \begin{center}
    \includegraphics[scale=0.34]{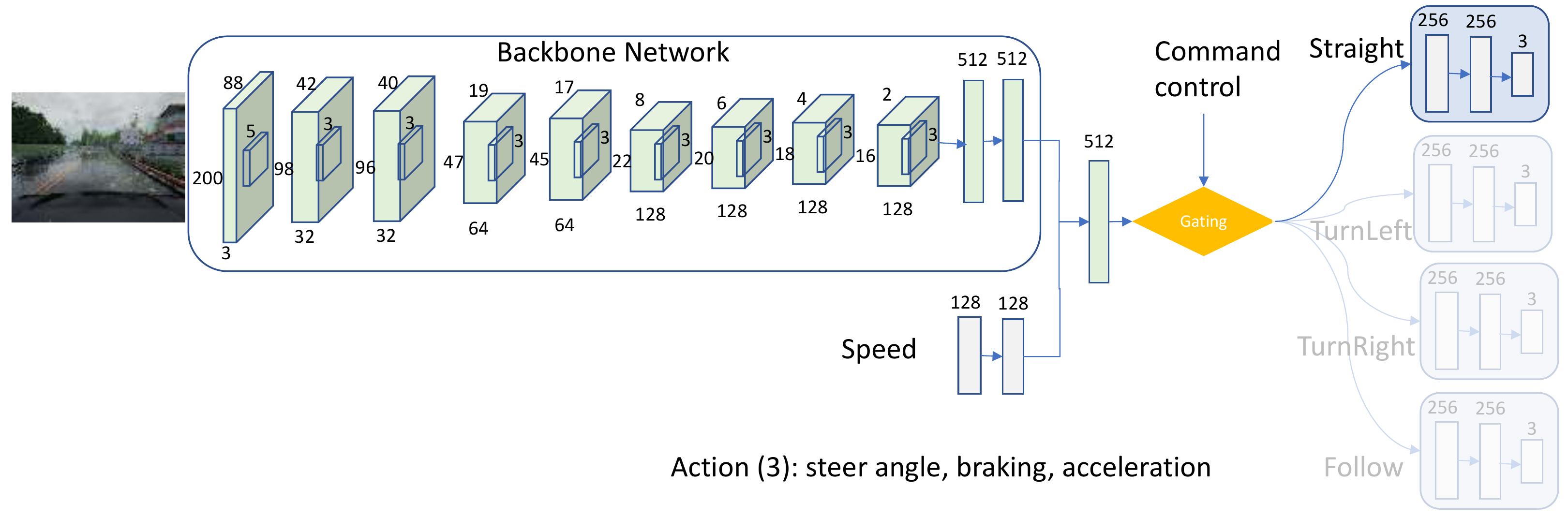}\vspace{-3mm}
            \caption{Actor Network Architecture of CIRL. The gating fuction selectively activates different branches to predict three actions for ``Straight", ``TurnLeft", ``TurnRight" and ``Follow" commands. }\vspace{-7mm} 
            \label{fig:actor}
        \end{center}
    \end{figure*}

\textbf{Reinforcement Learning for Autonomous Driving.}
Reinforcement learning learns by a trial-and-error fashion, and does not require explicit supervision from human. Deep-RL or RL algorithm has been applied to a wide variety of tasks, such as object recognition~\cite{liang2017deep,jie2016tree,han2018reinforcement,cao2017attention,liang2017recurrent}, computer games~\cite{mnih2015human}, robot locomotion~\cite{endo2008learning}, scene navigation~\cite{zhu2017target} and autonomous driving in the simulators~\cite{abbeel2007application,shalev2016safe,you2017virtual}. The most critical challenges in real-world applications are the high-dimensional large-scale continuous action space. Learning an optimal policy using such exhaustive exploration is prone to be very time-consuming and easy to fall into local optimum after many episodes. It is thus desirable to find a feasible action space that can help speed up the exploration. Our CIRL addresses this issue by leveraging learned experiences by imitation learning to guide the reinforcement driving agent. 

There exists some prior works also investigated the power of imitation learning. Generative Adversarial Imitation Learning (GAIL~\cite{ho2016generative}) builds a generative model, which is a stochastic policy that produces similar behaviors to the expert
demonstrations. InfoGAIL~\cite{li2017infogail} extends GAIL into a policy where low-level actions can be controlled through more abstract, high-level latent variables. The most similar work to ours are DQfD~\cite{hester2017learning}, ~\cite{latzke2006imitative} and DDPGfD~\cite{vevcerik2017leveraging}, which combines Deep Q Networks (DQN) with learning from demonstrations. However, DQfD is restricted to domains
with discrete action spaces, DQfD,~\cite{latzke2006imitative} and DDPGfD are not applicable for autonomous driving with significant different actor-critics, action spaces and reward definitions. Moreover, different with DDPGfD that loads the demonstration transitions into the replay buffer, we directly use the knowledge from demonstrations to guide the reinforcement explorations by initializing actor networks with pretrained model parameters via imitation learning. Our experiments show our strategy is particular better and more efficient than DDPGfD when applied to the autonomous driving simulator.

\section{Controllable Imitative Reinforcement Learning}

We illustrate the whole architecture of our CIRL method. To resolve the sample inefficiency issue in applying RL to complex tasks, our CIRL adopts an imitation stage and a reinforcement learning stage. First, given a set of human driving videos, we first use the supervised ground truth deterministic actions to pretrain the network. The command gating mechanism is incorporated to endow the model controllable capability for a central planner or drivers' intent. Second, to further enhance the policy with better generalization and robustness, the reinforcement learning optimization is employed to boost the ability of actor network. We first initialize the actor network with pretrained weights from the imitation stage, and then optimize it via the reward module by interacting with the simulator. Due to its superior performance on exploring continuous action space, we use the Deep Deterministic Policy Gradient (DDPG) as the RL optimization. Benefiting from the use of human driving demonstrations for initializing the actor network, the sample complexity can be significantly reduced to enable the learning within the equivalent of hours of exploration and interaction with the environment.

\subsection{Controllable Imitation Learning}

Given $N$ human driving video sequences $v_i, i\in (1,\dots, N)$ with the observation frame $I_{i,t}$, control command $c_{i,t}$, speed $s_{i,t}$, action $\mathbf{a}_{i,t}$ at each time step $t$, we can learn a deterministic policy network $F$ via the basic imitation learning to mimic the human experts. Detailed network architecture of $F$ is presented in Fig.~\ref{fig:actor}. The control command $c_{i,t}$ is introduced to handle the complex scenarios where the subsequent actions also depend on the driver's intent in addition to the observation~\cite{codevilla2017end}. The
action space $\mathbf{a}_{i,t}$ contains three continuous actions, that is steering angle $a^s_{i,t}$, acceleration $a^a_{i,t}$, and braking action $a^b_{i,t}$. The command $c_{i,t}$ is a categorical variable that control the selective branch activation via the gating function $G(c_{i,t})$, where $c_{i,t}$ can be one of four different commands, i.e. follow the lane (Follow), drive straight at the next intersection (Straight), turn left at the next intersection (TurnLeft), and turn right at the next intersection (TurnRight). Four policy branches are specifically learned to encode the distinct hidden knowledge for each case and thus selectively used for action prediction. 
The gating function $G$ is an internal direction indicator from the system. The controllable imitation learning objective is to minimize the parameters $\theta^I$ of the policy network $F^I$, defined as:
\begin{equation}
\min_{\theta^I} \sum_i^N\sum_t^{T_i}\mathcal{L}(F(I_{i,t},G(c_{i,t}), s_{i,t}), \mathbf{a}_{i,t}),
\label{eq:imitation}
\end{equation}
where the loss function $L$ is defined as the weighted summations of L2 losses for three predicted actions $\mathbf{\hat{a}}_{i,t}$:

\begin{equation}
\mathcal{L}(\mathbf{\hat{a}}_{i,t}, \mathbf{a_{i,t}}) = ||\hat{a}^s_{i,t}-a^s_{i,t}||^2 + ||\hat{a}^a_{i,t}-a^a_{i,t}||^2 + ||\hat{a}^b_{i,t}-a^b_{i,t}||^2,
\end{equation}
For fair comparison between our CIRL and imitation learning, we use the same  experiment setting as~\cite{dosovitskiy2017carla} to verify the effectiveness of boosting driving policies by our imitative reinforcement learning. The sensory inputs are images from a forward-facing camera, speed measurements from the simulator and control commands generated by the navigation planner.

\begin{figure*}[!tp]
        \begin{center}
     \includegraphics[scale=0.34]{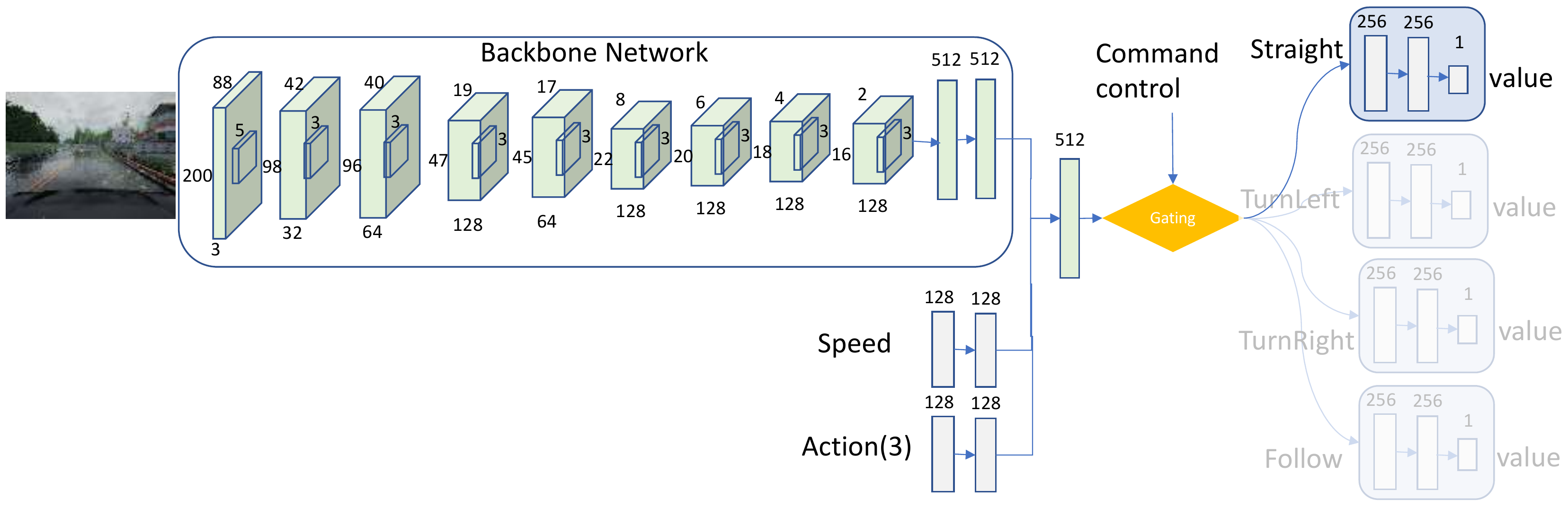}\vspace{-3mm}
            \caption{Critic Network Architecture of CIRL. The action outputs from actor network are fed into critic network to obtained the estimated value.}\vspace{-5mm}
            \label{fig:critic}
        \end{center}
    \end{figure*}
    
\subsection{Imitative Reinforcement Learning}

Our CIRL uses the policy network $F$ pretrained from conditional imitation learning to boost the sample efficiency of reinforcement learning to obtain more general and robust policies. We first present the underlying optimization techniques and then the reward designs.

\noindent\textbf{Markov Decision Process.} By interacting with the car simulator, the driving agent can be optimized based on a reward signal provided by the environment, with no human driving intervention, which can be defined as a Markov Decision Processes (MDPs)~\cite{sutton1998introduction}. In the autonomous driving scenario, the MDP is defined by a tuple of $<I,C,S,A,R,P, \lambda>$, which consists of a set of states $O$ defined with observed frames $I$, speeds $S$, control command $C$, a set of actions $A$, a reward function $R(o,\mathbf{a})$, a transition function
$P(o'|o, \mathbf{a})$, and a discount factor $\gamma$. In each state $o = <I, c, s>\in O$, the agent takes an action $\mathbf{a}\in A$. After taking this action and interacting with the environment, the agent receives a reward $R(o,\mathbf{a})$ and reaches a new state $o'$ depending on the probability distribution $P(o'|o, \mathbf{a})$. To make the driving policies more realistic, we focus on the goal-directed navigation, that is, the vehicle has to reach a predetermined goal along the path generated by the topological planner. The new observation $o'$ is thus updated by the simulator observation and a sequence of commands towards the goal. The episode is terminated when the vehicle reaches the goal, when the vehicle collides with an obstacle, or when a time budget is exhausted.

A deterministic and stationary policy $\pi$ specifies which action the agent will take given each state. The goal of the driving agent is to find the policy $\pi$ that maps states to actions that maximizes the expected discounted total reward. It can be thus learned by using a action value function: $Q^\pi(o,\mathbf{a}) = \mathbf{E}^\pi[\sum_{t=0}^{+\infty}\gamma^tR(o_t, \mathbf{a}_t)]$, where $\mathbf{E}^\pi$ is the expectation over the distribution of the admissible trajectories $(o_0, \mathbf{a}_0, \dots, o_t, \mathbf{a}_t)$ by executing the policy $\pi$ sequentially over the time episodes.

\noindent\textbf{Imitative Deep Deterministic Policy Gradient.} Since the autonomous driving system needs to predict continuous actions (steer angles, braking, and acceleration), we resort to the actor-critic approach for continuous control problem, and both actor and critic are parametrized by deep networks. Denoting the parameters of the policy network as $\theta$, and $\mu$ as the initial state distribution, the actor-critic approach aims to maximize a mean value $J(\theta) =\mathbf{E}_{o\sim \mu}[Q^{(\pi\dot|\theta)}(o,\pi(o|\theta))]$ in which $\theta$ can be updated via gradient descent as: $\theta + \alpha\nabla_\theta J(\theta)\rightarrow \theta$. In this work, we employ the Deep Deterministic Policy Gradient~\cite{lillicrap2015continuous} due to its promising performance on continuous control problem, which directly uses the gradient of Q-function with respect to the action for policy training. A policy network $F^\pi$ (actor) with parameters $\theta^\pi$ and an action-value function network $F^Q$ (critic) with parameters $\theta^Q$ are jointly optimized. The detailed network architectures of $F^\pi$ and $F^Q$ are presented in Fig.~\ref{fig:actor} and Fig.~\ref{fig:critic}.   
    
Different from the conventional DDPG that randomly initializes the $\theta^\pi$, our imitative DDPG proposes to load the pretrained $\theta^I$ in Eq.(\ref{eq:imitation}) via the imitation learning into $\theta^\pi$, obtaining a new $\bar{\theta^\pi}$ as the parameter initialization. It thus enables to produce reliable new transitions $e = (o, \mathbf{a}, r=R(o,\mathbf{a}), o'\sim P(\dot|o,\mathbf{a}))$ by acting based on $\mathbf{a}=\pi(o|\bar{\theta^\pi}) + \mathcal{N}$ where $\mathcal{N}\sim \text{OU}(\mu,\sigma^2)$ is a random process allowing action exploration. $\text{OU}(\cdot)$ denotes the Ornstein-Uhlenbeck process. Such further noisy exploration ensure that the agent’s behavior does not converge prematurely to a local optimum. The key advantage of our imitative DDPG lies in better initialized exploration starting points by learning from human expects, which can help significantly reduce the exhaustive exploration in the early stage of DDPG that may cost a few days, as discussed in previous works~\cite{plappert2017parameter}. Starting from a better state, the random action exploration allows RL to further refine actions according to the feedbacks from the simulator and results in more general and robust driving policies. The critic network is optimized by the one-step off-policy evaluation:
\begin{equation}
\mathcal{L}(\theta^Q) = \mathbf{E}_{(o,\mathbf{a}, r, o')\sim D}[R - Q(o,\mathbf{a} | \theta^Q)]^2,
\end{equation}
where $D$ is a distribution over transitions $e$ in the replay buffer and the one-step return $R = r + \gamma Q'((o', \pi'(o')|\bar{\theta^\pi}')|{\theta^Q}')$. $\bar{\theta^\pi}'$ and ${\theta^Q}'$ are parameters of corresponding target networks of $F^\pi$ and $F^Q$, which are used to stabilize the learning. On the other hand, the actor network is further updated from the starting state from the controllable imitative learning: 
\begin{equation}
\nabla_{\bar{\theta^\pi}}J(\bar{\theta^\pi})\approx\mathbf{E}_{o,\mathbf{a}\sim D}[\nabla_\mathbf{a}Q(o,\mathbf{a}|\theta^Q)_{|\mathbf{a}=\pi(o,\theta^Q)}\nabla_{\theta_\pi}\pi(o|\bar{\theta^\pi})].
\end{equation}

\begin{figure*}[!tp]
        \begin{center}
     \includegraphics[scale=0.46]{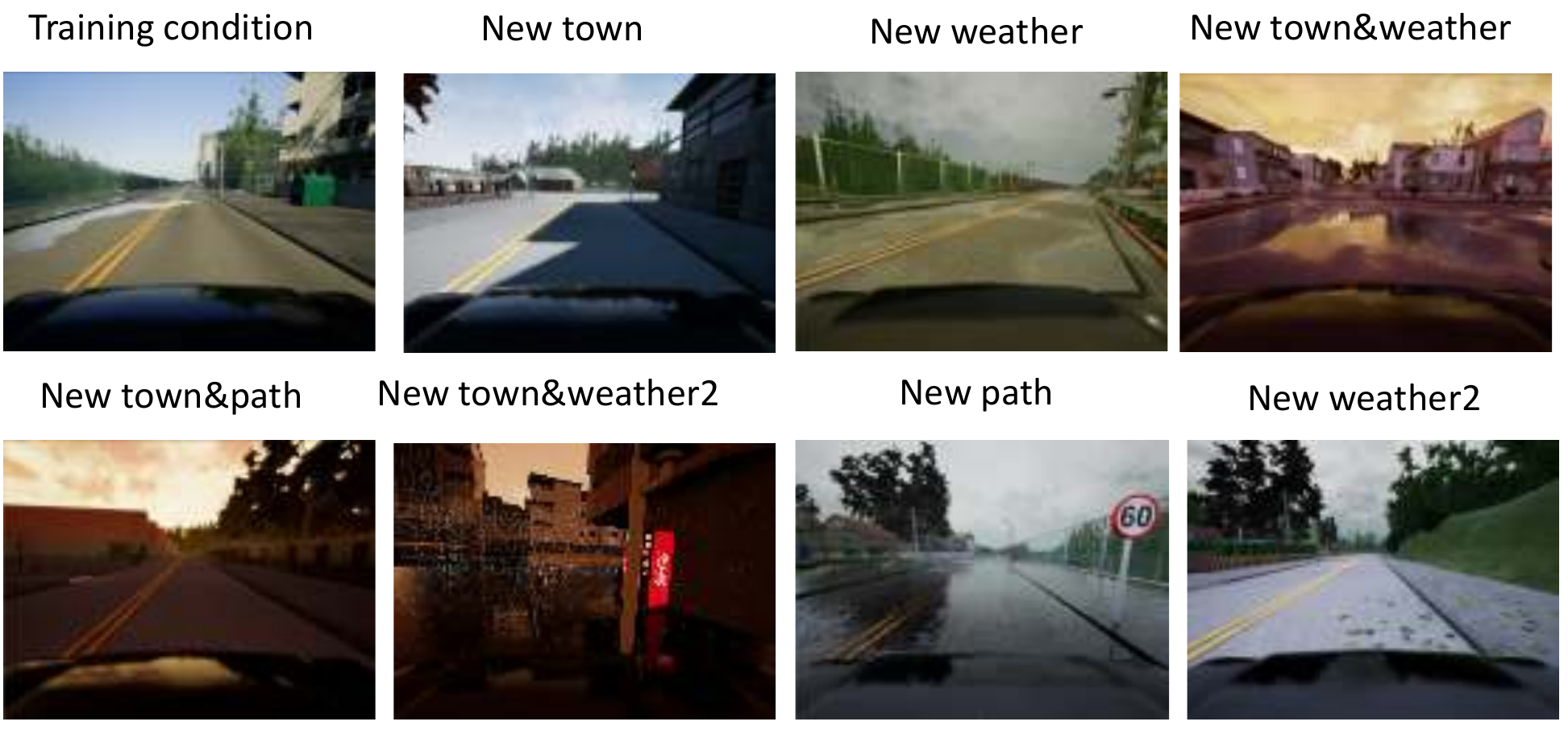}\vspace{-3mm}
            \caption{Example observations of different environment settings. Training condition is used for training while the rest settings are used for testing. Besides the settings (first row) evaluated in~\cite{dosovitskiy2017carla}, this work further validates the generalization capability of the model on four new settings (second row).}\vspace{-6mm}
            \label{fig:environments}
        \end{center}
    \end{figure*}
    
\noindent\textbf{Reward Module.} Another contribution of our CIRL is our reward module tailored for the autonomous driving scenario. The reward is a sum of five terms according to the measurements from simulator: negatively weighted abnormal steer angles $r_s$, positively weighted speed $r_v$ in km/h, and negatively weighted collision damage $r_d$, overlap with the sidewalk $r_r$, and overlap with the opposite lane $r_o$. The rewards are computed according to the simulator measurements after taking actions over the agent. First, the reward $r_s$ for abnormal steer-angles w.r.t each command control is defined as:
\begin{equation}
     r_s(c) =
  \begin{cases}
    -15       & \quad \text{if } s \text{ is in opposite direction with c for TurnLeft and TurnRight}\\
    -20  & \quad \text{if } |s| > 0.2 \text{, c for Straight.}
  \end{cases}
\end{equation}

Second, the reward $r_v$ for speed measurements after performing actions on the simulator with respect to each common control is defined as:

\begin{equation}
     r_v(c) =
  \begin{cases}
    \min(25,v)      & \quad \text{if } \text{c for Follow}\\
    \min(35,v)  & \quad \text{if }\text{c for Straight}\\
    v & \quad \text{if } v \leq 20 \text{, c for TurnLeft and TurnRight}\\
    40 - v & \quad \text{if } v > 20 \text{, c for TurnLeft and TurnRight}
  \end{cases}
\end{equation}

Finally, the $r_r$ and $r_o$ are both set as -100 for having overlapping with the sidewalk and opposite lane, respectively. The collision damage $r_d$ is as -100 for collision with other vehicles and pedestrians and as -50 for other things (e.g. trees and poles). The final reward $r$ conditioning on different command controls is computed as:
\begin{equation}
   r = R(o,\mathbf{a}) = r_s(c) + r_v(c) + r_r + r_o + r_d.
\end{equation}
Note that exact penalty values are applied for all experiments in our benchmark according to their specific limitations, such as speeds and angles~\cite{dosovitskiy2017carla}.
\section{Experiments}

\subsection{Experiment Settings}

\noindent\textbf{Evaluation benchmark.} We conduct extensive experiments on the recently release CARLA car simulator benchmark~\cite{dosovitskiy2017carla} because of its superior high-fidelity simulated environment and open-source accessibility, compared to other simulators. A large variety of assets were produced for CARLA, including cars and pedestrians. CARLA provides two towns: Town 1 and Town 2. For fair comparison with other state-of-the-art policy learning methods~\cite{dosovitskiy2017carla,codevilla2017end}, Town 1 is used for training and Town 2 exclusively for testing, as illustrated in Fig.~\ref{fig:environments}. The weather conditions are organized in three groups, including Training Weather set, New Weather set and New Weather2 set. Training Weather set is used for training, containing clear day, clear sunset, daytime rain, and daytime after rain. New Weather set and New Weather2 set are never used during training and for testing the generalization. New Weather set includes cloudy daytime and soft rain at sunset, and New Weather2 set includes cloudy noon, midrainy noon,  cloudy sunset, hardrain sunset. Besides three test settings evaluated in~\cite{dosovitskiy2017carla}, we further evaluate four new settings for more paths in Town 2, New weather2 set as shown in the first row in Fig.~\ref{fig:environments}. 

\begin{figure*}[!tp]
        \begin{center}
     \includegraphics[scale=0.42]{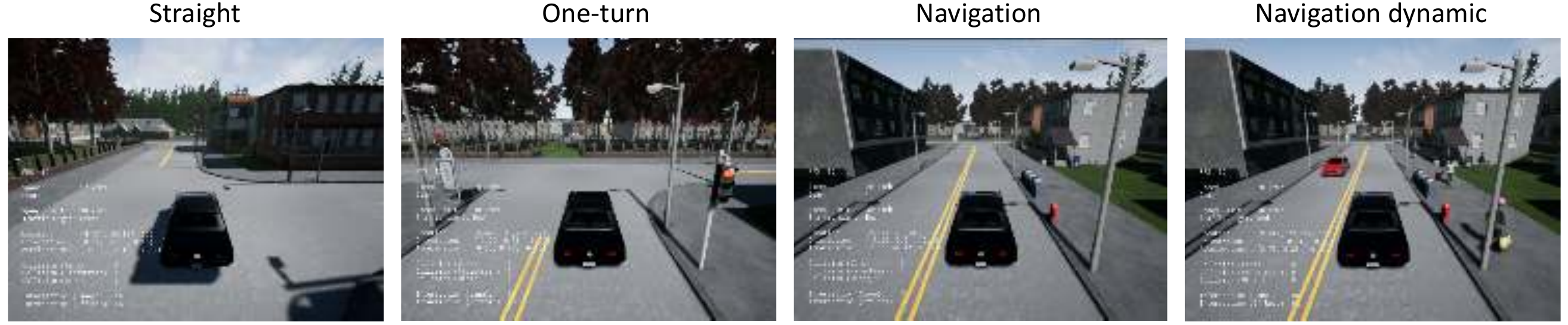}\vspace{-3mm}
            \caption{Illustrated observations of four different tasks in the bird view.}\vspace{-7mm}
            \label{fig:tasks}
        \end{center}
    \end{figure*}

\noindent\textbf{State-of-the-art pipelines.} We compare our CIRL model with three state-of-the-art pipelines in CARLA benchmark, that is modular pipeline (MP)~\cite{dosovitskiy2017carla}, imitation learning (IL)~\cite{dosovitskiy2017carla}, and reinforcement learning (RL)~\cite{dosovitskiy2017carla}, and fairly compete with them on four increasingly difficult driving tasks, i.e. Straight, One turn, Navigation and Navigation with dynamic obstacles, illustrated in Fig.~\ref{fig:tasks}. Particularly, the baseline MP~\cite{dosovitskiy2017carla} decomposes the driving task into the following subsystems including perception, planning and continuous control, and its local planning resorts to completely rule-based predefined policies that are completely dependent on the scene layout estimated by the perception module. The baseline IL~\cite{dosovitskiy2017carla} takes the images from a forward-facing camera and command controls as inputs, and directly trains the model via supervised learning using human driving videos. Note that for fair comparison, we adopt the same network architecture and settings with their model during the controllable imitation stage. RL~\cite{dosovitskiy2017carla} is also a deep reinforcement learning pipeline that uses the asynchronous advantage
actor-critic (A3C) algorithm~\cite{mnih2016asynchronous}. Different from their used five reward terms, we empirically remove the distance rewards traveled towards the goal since the way-points used for estimating distances are too sparse to give valid feedbacks during exploration. In addition, we propose to use controllable abnormal steer-angle rewards to penalize the unexpected angle predictions.

Note that for all methods, one same agent is used on all four tasks and cannot be fine-tuned separately for each scenario. The tasks are set up as goal-directed navigation: an agent is randomly initialized somewhere in town and has to reach a destination point. For each combination of a task, a town, and a weather set, the paths are carried out over 25 episodes. In each episode, the target of driving agent is to reach a given goal location. An episode is considered successful if the agent reaches the goal within a time budget, which is set to reach the goal along the optimal path at a speed of 10 km/h.

\noindent\textbf{Implementation settings.} During the controllable imitation stage, to fairly demonstrate the effectiveness of our imitative reinforcement learning, we use the exact same experiment settings in~\cite{codevilla2017end} for pre-training actor network. 14 hours of driving data collected from CARLA are used for training  and the network was trained using the Adam optimizer. Further details are referred in~\cite{codevilla2017end}.

During the imitative reinforcement learning stage, in terms of OU exploration parameters, we empirically set $\mu$ as 0, 0.15, and 0.5  and $\sigma$ as 0.02, 0.05, 0 for steer-angle, speed and braking actions, respectively. The discount factor $\gamma$ is set as 0.9. The initial learning rate of actor network is set as 0.00001 since it uses the shared weights from controllable imitation learning and the learning rate of critic network is set as 0.001. Learning and exploration rate are linearly decreased to zero over the course of training. The actor-critic networks are trained with 0.3 millions of simulation steps for roughly 12 hours of non-stop driving at 10 frames per second. In contrast, existing reinforcement learning approach provided in ~\cite{dosovitskiy2017carla} requires 10 millions of simulation steps corresponding to roughly 12 days of non-stop driving with 10 parallel actor threads. Our CIRL can obtain high percentage of successfully completed episodes after several hours with good sample efficiency, benefiting from a good exploration start boosted by the controllable imitation stage. The proposed method is implemented on TensorFlow framework. All models are trained on four NVIDIA GeForce GTX1080  GPUs.

\begin{table}[t]
\centering
\small
\caption{ Quantitative comparison with other state-of-the-art autonomous driving systems on four goal-directed navigation tasks. The table reports the percentage (\%) of successfully completed episodes in each condition. Higher
is better. The tested methods are: modular pipeline (MP)~\cite{dosovitskiy2017carla}, imitation learning (IL)~\cite{dosovitskiy2017carla}, and reinforcement
learning (RL)~\cite{dosovitskiy2017carla} and our CIRL model.}\vspace{-3mm}
\tabcolsep 0.015in 
\begin{tabular}{c|cccc|cccc|cccc|cccc|}
\toprule[0.1pt]
\multirow{2}{*}{Task}   & \multicolumn{4}{c}{Training conditions} & \multicolumn{4}{c}{New town} & \multicolumn{4}{c}{New weather} & \multicolumn{4}{c}{New town/weather} \\ & MP & IL & RL & CIRL & MP & IL & RL & CIRL & MP & IL & RL & CIRL & MP & IL & RL & CIRL\\ \hline                     
Straight & 98 & 95 & 89  & \textbf{98} &  92  & 97  & 74 & \textbf{100} & 100 & 98 & 86 & \textbf{100} & 50 & 80 & 68 & \textbf{98} \\
One turn & 82 & 89 & 34 & \textbf{97} & 61  & 59  & 12 & \textbf{71} & \textbf{95} & 90 & 16 & {94} & 50 & 48 & 20 & \textbf{82} \\   
Navigation & 80 & 86 & 14 & \textbf{93} & 24  & 40  & 3 & \textbf{53} & \textbf{94} & 84 & 2 & {86} & 47 & 44 & 6 & \textbf{68}\\
Nav. dynamic & 77 & \textbf{83} & 7 & {82} & 24  & 38 & 2 & \textbf{41} & \textbf{89} & 82 & 2 & {80} & 44 & 42 & 4 & \textbf{62}\\ 
\toprule[0.7pt]
\end{tabular}\vspace{-6mm}
\label{tab: results_origin}
\end{table}

\subsection{Comparisons with state-of-the-arts}
    
Table~\ref{tab: results_origin} reports the comparisons with the state-of-the-art pipelines on CARLA benchmarks in terms of the percentage of successfully completed episodes under four different conditions. All results of MP, IL and RL were reported from~\cite{dosovitskiy2017carla}. For ``Training conditions" task, the models are tested on the combination of Town 1, Training Weather setting which has different starting and target locations under the same general environment and conditions with the training stage. The rest test settings are conducted for evaluating more aggressive generalization, that is, adaption to the previously unseen Town 2 and to previously unencountered weather from the New Weather and New Weather2.

\begin{figure*}[!tp]
        \begin{center}
     \includegraphics[scale=0.43]{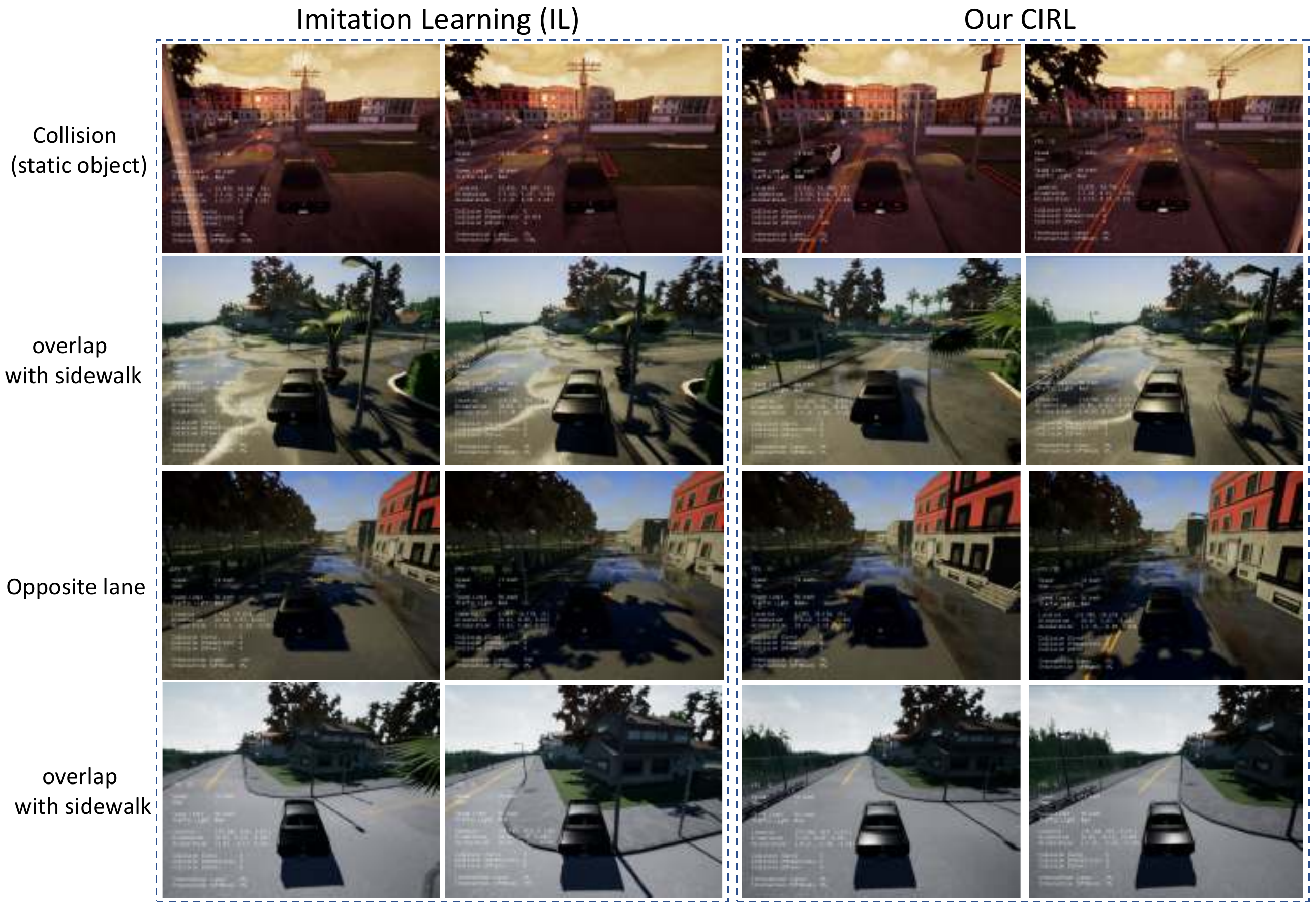}\vspace{-3mm}
            \caption{Visualization comparisons between the imitation learning baseline~\cite{dosovitskiy2017carla} and our CIRL model. We illustrate some driving cases for straight and one-turn tasks, and show the IL baseline fails with some types of infractions (e.g. collision with static object, more than 30\% overlap with Sidewalk, in opposite lane) while our CIRL successfully completes the goal-oriented tasks. For each case, two consecutive frames are shown.}\vspace{-6mm}
            \label{fig:visualresult}
        \end{center}
    \end{figure*}

We can observe that our CIRL substantially outperforms all baseline methods under all conditions, especially better than their RL baseline. Furthermore, our CIRL shows superior generalization capabilities in the rest three unseen setting (e.g. unseen new town), which obtains not perfect results but considerably better performance over other methods, e.g. 71\% of our CIRL vs. 59\% and 12\% of IL and RL, respectively. More qualitative results are shown in Fig.~\ref{fig:visualresult}, which provides some infraction examples that the IL model suffers from and our CIRL successfully avoids. 

It is also interesting that both learning-based methods (IL and our CIRL) achieve comparable and better performances than the modular pipeline, although MP adopted the  sophisticated perception steps (segmentation and classification) to identify key cues in the environment and used manually rule-based policies. One exception is that the modular pipeline performs better under the ``New weather" condition than that of the training conditions, and both IL and CIRL are slightly inferior to it. But MP's results perform bad on navigation task and considerably decrease on all tasks in unseen ``New town" and ``New town/weather" conditions. The reason is that MP heavily depends on the perception stage that fails systematically under complex weather conditions in the context of a new environment, and rule-based policies that may fail for long-range goal-driven navigation. We can conclude that MP is more fragile to unseen environments than the end-to-end learning based models since the perception part itself is difficult and hard to adapt across diverse unknown scenes.

\begin{table}[t]
\centering
\small
\caption{The percentage (\%) of successfully completed episodes of our CIRL on four new settings for further evaluating generalization.}\vspace{-3mm}
\tabcolsep 0.03in 
\begin{tabular}{c|c|c|c|ccc|cccccc|cccc|}
\toprule[0.1pt]
\multirow{1}{*}{Task}   & \multicolumn{1}{c}{New town/path2} & \multicolumn{1}{c}{New town/weather2} & \multicolumn{1}{c}{New path} & \multicolumn{1}{c}{New weather2} \\ 
\hline               
Navigation  &50 & 58 & 95 & 87\\
Nav. dynamic & 38 & 47 & 87 & 86\\
\toprule[0.7pt]
\end{tabular}\vspace{-4mm}
\label{tab: results_new}
\end{table}

\begin{table}[t]
\centering
\small
\caption{The percentage (\%) of successfully completed episodes of our CIRL under different weather conditions for the navigation tasks in training town and new town.}\scriptsize\vspace{-3mm}
\begin{tabular}{c|c|c|c|c|c}
\toprule[1pt]
   Navigation task & CloudyNoon & MidRainyNoon & CloudySunset & WetCloudySunset & HardRainSunset\\ \hline 
   CIRL (Town 1) & 92 & 96 & 96 & 64 &56\\
   CIRL (New Town)  & 95 & 52 & 85 & 90 & 5\\
\toprule[0.7pt]
\end{tabular}\vspace{-6mm}
\label{tab: weather}
\end{table}

On the other hand, the  conventional reinforcement learning~\cite{dosovitskiy2017carla} performs significantly worse than all other methods, even with considerably more training time: 12 days of driving in the simulator. The reason is that RL itelf is well known to be brittle~\cite{henderson2017deep} and needs very time-consuming exploration to get reasonable results. Rather than video games in Atari~\cite{mnih2015human} and maze navigation~\cite{dosovitskiy2016learning}, the real-world tasks like self-driving require complex decision making to exploit visual cues, leading to severe sample inefficiency and unfeasible parameter search. 

In contrast, the proposed CIRL effectively benefits from both merits of imitation learning (i.e. fast convergence) and traditional reinforcement learning (i.e. robust long-term decision making). Our CIRL that enhances the policies by only rough 12 hours of driving explorations in car simulator can achieve significant better performances on all tasks than the best MP and IL methods. Different from previous RL models that conduct too much random and meaningless explorations in the beginning, the actor network in our CIRL can start the exploration in a good and reasonable point by transferring knowledge from the first controllable imitation stage. The reward feedbacks by driving and interacting with complex dynamics in the simulator can further facilitate the policy learning with better robustness and generalization capability. 


\subsection{Generalization capability} The exact driving trajectories during training cannot be repeated during testing. Therefore performing perfectly on CARLA benchmark requires robust generalization, which is challenging for existing deep learning methods. As reported in Table~\ref{tab: results_origin}, it is obvious that all methods perform closely to those in ``Training conditions" under the ``New weather" setting. However, their performances dramatically drop on the ``New town" settings. For example, on the most challenging navigation task ``Nav.dynamic" in the New town/weather setting, previous best MP and IL methods obtain only 44\% and 42\% complete success episodes compared to 62\% of our CIRL. In general, our CIRL shows much better generalization capabilities over other methods, but still needs further improvements. 

Besides the previous two types of generalization (i.e. unseen weather conditions and unseen new town), we further conduct more experiments on two another new conditions (i.e. more path trajectories and the New weather2 set) on two most difficult tasks to further evaluate more general cases, resulting in four new settings in Table~\ref{tab: results_new}. We can see that our model shows reasonably robust and good performance on different navigation paths and weather set. Adapting our CIRL to navigate in unseen towns can be improved by training in wider range of different scenes. This further demonstrates well the advantages of integrating together the controllable imitation learning and DDPG algorithm into boosting driving policies towards more challenging tasks.

\begin{figure*}[!tp]
        \begin{center}
     \includegraphics[scale=0.4]{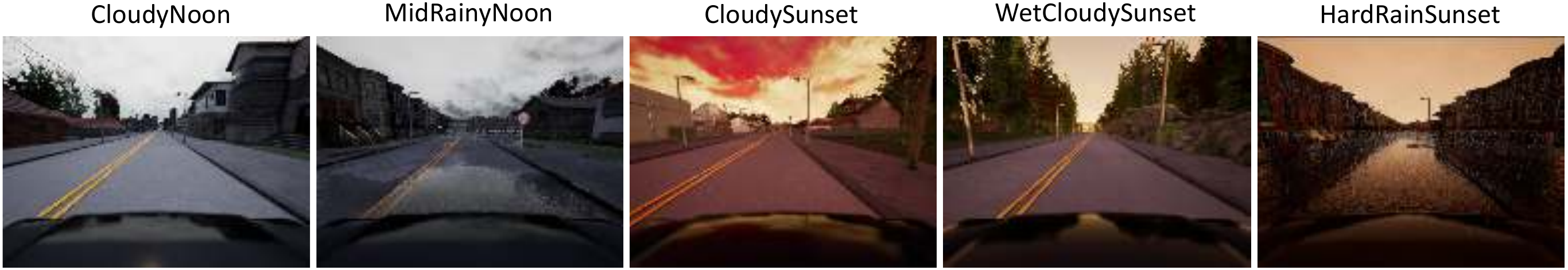}\vspace{-3mm}
            \caption{Example observations under distinct weather conditions. Better viewed in zoom.}\vspace{-10mm}
            \label{fig:weather}
        \end{center}
    \end{figure*}

We also extensively dive into the affects of different weather conditions on driving generalization capability, as reported in Table~\ref{tab: weather}. Driving behaviors under five weather conditions with distinct levels of difficulties are evaluated on both seen town and unseen town. We can observe promising results obtained under weathers with good visibility, such as CloudyNoon, CloudySunset. But regarding to more challenging rainy weathers, the model obtains very low successfully completed rates. One of main reasons is that the road and surrounding dynamics are extremely hard to be perceived as a result of heavy rains, as shown in Fig.~\ref{fig:weather}.

\begin{table}[t]
\centering
\caption{Ablation studies on one-turn task on four different settings.}\footnotesize\vspace{-3mm}
\begin{tabular}{c|c|c|c|cc}
\toprule[1pt]
   Method (one-turn) & Training conditions & New town & New weather & New town/weather\\ \hline 
   CIRL w/o steer reward  & 91         & 65 & 96 & 76  \\
   CIRL w/ add replay & 96  & 71 & 94 & 82\\
   CIRL more simulation steps & 95  & 68 & 98 & 80\\
   \hline
   {Our CIRL} & {97}  & {71} & {94} & {82}\\
\toprule[0.7pt]
\end{tabular}\vspace{-4mm}
\label{tab: ablation}
\end{table}

\begin{table}[t]
\centering
\caption{Results on comma.ai dataset in terms of mean absolute error (MAE).}\label{tab:realscene}\footnotesize\vspace{-3mm}
\begin{tabular}{c|c|c|c|cc}
\toprule
\textbf{Model} & PilotNet~\cite{bojarski2016end} & CIRL (CARLA) & CIRL from scrach & CIRL finetuning & \\ \hline
\textbf{Steer-angle MAE}   & 1.208  & 2.939  & 1.186 & \textbf{1.168} \\
\bottomrule
\end{tabular}\vspace{-4mm}
\end{table}

\subsection{Comparisons on real scenes}

We report results of applying our CIRL trained on CARLA into real scenes in Table~\ref{tab:realscene} on  Comma.ai~\cite{santana2016learning} dataset. To finetune on Comma.ai, we use pretrained network parameters before direction branches on CARLA and initialize 3 stacked fc-layers (256,256,1) to predict one steer angle. The learning rate is set to 1e-3. We train 18 epochs and batch size is 256. `` CIRL (CARLA)" denotes directly applying model trained on CARLA into prediction in real scenes. We can see that finetuning pretrained CIRL model on comma.ai (``CIRL finetuning") outperforms the baseline PilotNet and ``CIRL from scratch" that is trained from scratch on Comma.ai. It verifies well that our CIRL model learned from the high-quality CARLA simulator can be easily transferred into real scenes to enhance driving policy learning for real autonomous vehicles.

\begin{table}[t]
\centering
\caption{Success rates on \emph{One Turn} task in New Town (i.e. validation town)}\label{tab:reward}\footnotesize\vspace{-3mm}
\begin{tabular}{c|c|c|c|c|c}
\hline
\textbf{Reward} & our\_reward & our\_reward$\times$10 & our\_reward/10 & w/o speed & w/o offroad\&coll\\ 
\hline
Old weather & \textbf{71\%} & 70\% & 52\% & 20\%          & 31\%\\
\hline
New Weather & \textbf{82\%} & 82\% & 68\% & 14\% & 28\%\\
\hline
\end{tabular}\vspace{-6mm}
\end{table}
    
\subsection{Ablation studies}
We also conduct comprehensive experiments to verify the effects of each key component of our model, as reported in Table~\ref{tab: ablation}. Experiments are conducted on the challenging one-turn task on four different environments.

\noindent\textbf{Different strategies of using demonstrations.} To validate the effectiveness of our imitative reinforcement learning, we compare our CIRL with DDPGfD~\cite{vevcerik2017leveraging} that performs learning from demonstrations for robotic manipulation problems. In contrast to our strategy of providing a better exploration start, DDPGfD instead loads the demonstration transitions into the replay buffer and keeps all transitions forever. We thus implement and incorporate the demonstrate replay buffer into our CIRL, and ``CIRL w/ add reply" denotes the results of this variant for running the same number of simulation steps with our CIRL. We can see there is no noticeable performance difference between ``CIRL w/ add reply" and our CIRL. It speaks well that the good starting point for exploration is already enough for learning reasonable policies in an efficient way. We also try the performance of pure DDPGfD on our task without using imitation learning to initialize the actor network, which is quite bad after several days of driving simulation due to the need of exhaustive exploration, we thus did not list their results. Note that for justifying the optimization step, we keep all experiments settings of all variants as same, e.g. reward design.

\noindent\textbf{The effect of abnormal steer-angle rewards.} Different from the reward terms in~\cite{dosovitskiy2017carla}, we propose to adopt specialized steer-angle rewards with respect to each command control. Our comparisons between ``CIRL w/o steer reward" and ``CIRL" further demonstrate the effectiveness of incorporating such rewards for stabilizing the action exploration by providing more explicit feedbacks.

\noindent\textbf{The effect of simulation step number.} One raised question for our CIRL is whether the performance can be further improved by performing RL policy learning with more simulation steps. ``CIRL more simulation steps" reports results of running CIRL model for 0.5 million steps. We find that no significant improvement in terms of percentages of completely success episodes can be obtained in unseen driving scenarios. This verifies our model can achieve good policies by efficient sample exploration with the acceptable computation cost. On the other hand, this may motivate us to further improve model capability from other aspects, such as exploring more environments and video dynamics to improve the generalization ability. 

\noindent\textbf{Reward function.} set scales of reward values following Coach RL framework\footnote{https://nervanasystems.github.io/coach/} used in CARLA environment. Ablation studies on different reward scales for all rewards are reported in Table~\ref{tab:reward}. We can observe that removing speed or offroad\&collision reward significantly decreases the success rate. Moreover, using 10x larger reward values obtains minor performance difference while 10x smaller rewards lead to worse results.

\section{Conclusion}

In this paper, we propose a novel CIRL model to address the challenging problem of vision-based autonomous driving in the high-fidelity car simulator. Our CIRL incorporates controllable imitation learning with DDPG policy learning to resolve the sample inefficiency issue that is well known in reinforcement learning research. Moreover, specialized steer-angle rewards are also designed to enhance the optimization of our policy networks based on controllable imitation learning. Our CIRL achieves the state-of-the-art driving performance on CARLA benchmark and surpasses the previous modular pipeline, imitation learning and reinforcement learning pipelines. It further demonstrates superior generalization capabilities on a variety of different environments and conditions.

\clearpage

\bibliographystyle{splncs}
\bibliography{egbib}
\end{document}